%% file: acl.tex
\pdfoutput=1

\documentclass[11pt]{article}

\usepackage{acl}
\usepackage{times}
\usepackage{latexsym}
\usepackage{booktabs}
\usepackage{algorithm}
\usepackage{algorithmic}
\usepackage{amsmath}
\usepackage{multirow}
\usepackage{makecell}
\usepackage[T1]{fontenc}

\usepackage[utf8]{inputenc}

\usepackage{microtype}

\title{Diversifying Neural Dialogue Generation via Negative Distillation}

\begin{document}

\author{Yiwei Li, Shaoxiong Feng, Bin Sun, Kan Li\\
  School of Computer Science, Beijing Institute of Technology \\
  \texttt{\{liyiwei,shaoxiongfeng,binsun,likan\}@bit.edu.cn}}

\maketitle
\input{abstract}
\input{introduction}
\input{method}
\input{experiment}
\input{related_work}
\input{conclusion}

\section*{Acknowledgements}
We would like to thank the anonymous reviewers for their constructive comments. This work is supported by Beijing Natural Science Foundation (No.4222037, L181010) and National Natural Science Foundation of China (No.61972035). Kan Li is the corresponding author.

\bibliography{anthology,custom}
\bibliographystyle{acl_natbib}

\input{appendix}

\end{document}

%% file: abstract.tex
\begin{abstract}
Generative dialogue models suffer badly from the generic response problem, limiting their applications to a few toy scenarios. 
Recently, an interesting approach, namely negative training, has been proposed to alleviate this problem by reminding the model not to generate high-frequency responses during training.
However, its performance is hindered by two issues, ignoring low-frequency but generic responses and bringing low-frequency but meaningless responses.
In this paper, we propose a novel negative training paradigm, called negative distillation, to keep the model away from the undesirable generic responses while avoiding the above problems. 
First, we introduce a negative teacher model that can produce query-wise generic responses, and then the student model is required to maximize the distance with multi-level negative knowledge.
Empirical results show that our method outperforms previous negative training methods significantly.\footnote{The code and preprocessed data are available at https://github.com/Yiwei98/dialogue-negative-distillation.}

\end{abstract}

%% file: introduction.tex
\section{Introduction}

In the past few years, data-driven response generation \citep{NoisyData-Vinyals-2015,shang-etal-2015-neural,vougiouklis-etal-2016-neural} has achieved impressive performance, drawing continuously increasing attention from academia and industry. 
Conventionally, with the guidance of maximum likelihood estimation (MLE), neural dialogue models are expected to maximize the probability of generating the corresponding reference given any query. 
Unfortunately, due to the many-to-one phenomenon (see Table~\ref{tb:intro}), a characteristic of the dialogue task \citep{csaky-etal-2019-improving}, these models are prone to produce safe but generic responses (e.g., \textit{I don't know} \citep{li-etal-2016-diversity}), which sets an obstacle for the generative dialogue system to be deployed widely. 
Some researchers tried to redesign the objective of models to meet the requirement of diverse responses instead of MLE, such as MMI \citep{li-etal-2016-diversity}, AdaLabel \citep{wang-etal-2021-diversifying}, and IAT \citep{zhou-etal-2021-learning}. 
Besides, several studies \citep{kulikov-etal-2019-importance,DBLP:conf/iclr/HoltzmanBDFC20} proposed more advanced decoding strategies to alleviate the problem of generic responses. 
Indeed, the above methods boost the diversity of responses by reminding the model what should be said.

\begin{table*}[h]
    \centering

    \begin{tabular}{l l l c}
    \toprule
         & Query & Response &  Entropy Score\\  \midrule
        1: & Oh, honey, you made a mistake.  & \textbf{I don't know} how to do it. & 8.61 \\ \midrule
        2: & Would you like regular car wash package? & \textbf{I don't know} what you mean.  & 8.75 \\ \midrule
        3: & I'm looking for the airport. & No, sorry. \textbf{I don't know}. & 10.69 \\
        & Can you tell me how to get there?  \\ \midrule
        4: & That's cathy. She is pretty, isn't she?  & Well, \textbf{I don't know}. & 12.14 \\     
        & She likes dancing. Ask her to dance.   \\ \midrule
        5: & It doesn't matter. You gotta find what she's  & \textbf{I don't know} . . .  & 6.82 \\
        & interested in and go with that. \\ 
    \bottomrule
    \end{tabular}
    \caption{The \textit{many-to-one} phenomenon in DailyDialog. All the above five queries have the same \textbf{I don't know}-like responses. The corresponding \textit{source entropy} \citep{csaky-etal-2019-improving} scores are much higher than the median score (0.92) of the whole training set. This phenomenon will lead to the generic response problem.}
    \label{tb:intro}
\end{table*}

However, inspired by negative training \citep{DBLP:conf/iccv/KimYYK19,ma-etal-2021-sent}, we argue that it is also necessary to tell the dialogue model what not to say. To alleviate the problem of generic responses, \citet{he-glass-2020-negative} negatively updates the parameters when identifying the high-frequency responses. \citet{li-etal-2020-dont} punishes the behaviors of generating repetitive or high-frequency tokens by using the unlikelihood objective \citep{DBLP:conf/iclr/WelleckKRDCW20}.

Although the negative-training based methods enhance the diversity of responses, there still exists two drawbacks: 
First, they regard high-frequency tokens or utterances as negative candidates. However, the high-frequency response problem is only a sub-problem of the generic response problem \citep{he-glass-2020-negative}. It means that the responses that are low-frequency but generic will escape from punishment. Even worse, we have observed that some generic responses followed by a low-frequency but meaningless subsequence can avoid being identified as high-frequency, which inevitably sacrifices the fluency of responses (see Analysis). 
Second, these methods ignore the implicit negative knowledge in neural networks that characterizes negative candidates at multiple levels. We contend that it is more effective to conduct negative training with richer information (e.g., hierarchical representation).

To tackle the above problems and further improve the diversity of responses, we propose a novel negative training paradigm called \textit{Negative Distillation} (ND). Conventional knowledge distillation (KD) \citep{DBLP:journals/corr/HintonVD15,jiao-etal-2020-tinybert} takes the teacher as a positive role model and induces the student to imitate. Differing from that, we train the teacher as a negative role model and remind the student to get rid of those bad behaviors.

Specifically, we first collect a negative training set by using a filtering method called \textit{Source Entropy} \citep{csaky-etal-2019-improving}. This filtering method can retrieve all \textit{many-to-one} cases of the raw dataset. Note that the ``one'' is usually a generic response. 
Then, we train a dialogue model on the above sub-set as the negative teacher. Given queries, the negative teacher can provide a set of negative candidates (i.e., generic and dull responses) that the student is prone to generate, which avoids the first drawback mentioned before. 
Therefore, the student obtains query-wise bad behaviors for \textit{Negative Distillation}. 
To conduct the negative update holistically, we design two negative objectives, including soft unlikelihood loss on the prediction layer and reverse square error on the intermediate layer.
In this way, the negative distillation fully exploits multi-level negative knowledge to force the student to generate non-generic responses.

Our contributions are summarized as follows:
\begin{itemize}

\item We propose a novel and effective negative training paradigm called \textit{Negative Distillation}. It constructs query-wise generic responses as the negative candidates.
\item We design two negative objectives to utilize multi-level information to further boost the performance of negative distillation. 
\item We perform extensive experiments and detailed analysis to verify the effectiveness of the negative distillation framework and the superiority compared with previous negative training methods.

\end{itemize}

%% file: method.tex
\section{Method}
In this section, we first introduce the negative teacher, then describe the negative distillation on the prediction layer and the intermediate layer, respectively, and finally present the progressive optimization objective. Algorithm~\ref{algorithm} shows the whole training details.

\subsection{Background}

\paragraph{Dialogue Generation with MLE} 
Take $Q=\{q_1, q_2,...,q_{T_{q}}\}$ and $R=\{r_1, r_2,...,r_{T_{r}}\}$ as the (\textit{query, response}) pair, where $T_{q}$ and $T_{r}$ represent the length of query and response, respectively. The generative dialogue model aims to learn a conditional probability distribution $p_{\theta}(R|Q)$. The maximum likelihood estimation (MLE) is usually used to train the model, which can also be expressed as minimizing the negative log-likelihood:

\begin{equation}
\mathcal{L}_{\mathrm{MLE}}=-\sum_{i=1}^{T_{r}} \log p_{\theta}\left(r_{i} \mid r_{<i}, Q \right).
\label{eq:mle}
\end{equation}
Considering one characteristic of the dialogue task, i.e., allowing the response to be varied, the \textit{many-to-one} phenomenon occurs in the dialogue corpora frequently. However, with the MLE-based training, this phenomenon will cause the model to produce generic responses.

\paragraph{Unlikelihood Training} Unlikelihood (UL) loss \citep{DBLP:conf/iclr/WelleckKRDCW20} is proposed for the model to address the problem of undesirable behaviors (e.g., repetitive or high-frequency tokens). It forces the model to minimize the probability of generating negative candidates, which is formulated as:
\begin{align}
\nonumber  \mathcal{L}_{\mathrm{UL}}=  -\sum_{i=1}^{T_{r}}  \sum_{r_{c} \in \mathcal{C}_{t}} \\
\log(1- p_{\theta}(r_{c}  \mid r_{<i}&, Q )), 
\end{align}
where $\mathcal{C}_{t}$ consists of negative candidates (e.g., overuse frequent words) that are also a sub-set of the vocabulary.

\paragraph{Knowledge Distillation} The traditional knowledge distillation (KD) usually transfers useful knowledge from a large and strong teacher network $T$ to a small student network $S$. The distillation loss is used to align the soften predictions of the teacher and the student, denoted as $f^{T}(x)$ and $f^{S}(x)$:

\begin{equation}
\mathcal{L}_{\mathrm{KD}}=\sum_{x \in \mathcal{D}} L\left(f^{T}(x), f^{S}(x)\right),
\end{equation}
where $L(\cdot)$ is a measurement function that calculates the distance of different probability distributions, $x$ is the input text, and $\mathcal{D}$ denotes the training set.

In this work, we replace the positive teacher in vanilla KD with a negative teacher, aiming to provide negative knowledge for the student to conduct negative training and avoid undesirable behaviors.

\subsection{Negative Teacher}

To improve the diversity of responses, the dialogue model should be told which responses are generic. 
For negative distillation, a negative teacher is required to produce possible generic responses given any query. 
In this work, we adopt the widely used Transformer \citep{Transformer-Vaswani-2017} as the underlying model for both teacher and student. 
We introduce the \textit{Source Entropy} filtering method \citep{csaky-etal-2019-improving} to identify and collect the many-to-one cases for the negative training set. The source entropy is defined as:
\begin{equation}
    H_{src}(r, \mathcal{D}) = -\sum_{(q_i, r)\in \mathcal{D}}p(q_i|r)\log{p(q_i|r)},
\label{eq:h}
\end{equation}
where $p(q_i|r)$ is the conditional probability calculated based on the relative frequency of (\textit{query, response}) pairs, $r$ is a response, $q_i$ is the query corresponding to the response $r$, and $\mathcal{D}$ represents the raw training set. A higher source entropy indicates that the response $r$ corresponds to more queries, i.e., the \textit{many-to-one} problem is serious. We select the top 50\% \footnote{Simply the same as \citet{akama-etal-2020-filtering}} dialogue pairs $(q,r)$ with a high source entropy as the negative training set $\mathcal{D}_N$, which contains a much higher proportion of generic responses than the raw training set. 

After that, we train the teacher $N$ on the negative training set $\mathcal{D}_N$ by Equation~\ref{eq:mle}. The teacher will naturally produce generic responses for any input query. More importantly, it will provide richer negative knowledge for the student, including soft logits in the prediction layer and implicit features in the intermediate layers.

\subsection{Negative Distillation}
In this section, we conduct the negative distillation for the student based on the multi-level negative knowledge. 

\paragraph{ND for Prediction Layer}\label{md:pred}

The soften logits in the prediction layer contain more information than the ground-truth labels, such as the similarity between labels \citep{wang-etal-2021-diversifying}. Therefore, conventional KD transfers knowledge by narrowing the gap between the probability distributions of the teacher $T$ and the student $S$:

\begin{align}
\nonumber  \mathcal{L}_{KD}  & =  -\sum_{i=1}^{T_{r}} \sum_{k=1}^{|\mathcal{V}|} p_{T} \left(r_{i}=k \mid r_{<i}, Q\right)  \\
 & \cdot \log p_{S}\left(r_{i}=k \mid r_{<i}, Q \right).
\end{align}
As for negative distillation, the extra knowledge in soften logits of the negative teacher reflects how to generate dull responses based on the input query. Therefore, we propose a soft unlikelihood loss to maximize the distance between the predictions of the negative teacher $N$ and the student $S$:

\begin{align}
\label{eq:pred}
\nonumber   \mathcal{L}_{pred}= & -\sum_{i=1}^{T_{r}} \sum_{k=1}^{|\mathcal{V}|} p_{N} \left(r_{i}=k \mid r_{<i}, Q \right) \\
& \cdot \log \left(1-p_{S}\left(r_{i}=k \mid r_{<i}, Q \right)\right),
\end{align}

where $p_{N}$ and $p_{S}$ are calculated by:
\begin{equation}
p^{i}=\frac{\exp \left(z_{i} / t\right)}{\sum_{j} \exp \left(z_{j} / t\right)},
\end{equation}
where $t$ is a temperature coefficient that is used to soften the probability distribution over words. 

It should be emphasized that previous negative training methods only use the high-frequency words or phrases with one-hot representation as the targets, which ignores the rich information existing in the soften logits (e.g., the generic words have similar probabilities). 
In the Analysis section, we demonstrates the superiority of soften logits compared with hard targets (i.e., one-hot representation).

\paragraph{ND for Intermediate Layer}

In addition to the output knowledge from the prediction layer, there is also some implicit knowledge embedded in the intermediate layers, such as hidden states and attention matrices. To keep the student away from undesirable behaviors (i.e., producing generic responses) more effectively, we further consider the above knowledge into negative distillation. Specifically, the distance between features of the negative teacher and the student should also be increased. In this work, we propose a new measurement function, called mean reverse square error (MRSE), to calculate this distance:

\begin{equation}
    \mathcal{L}_{MRSE}(\boldsymbol{A}, \boldsymbol{B}) = \frac{1}{n} \sum^{n}_{i=1} \exp^{-SE(\boldsymbol{A}_{i}, \boldsymbol{B}_{i})},
\end{equation}
where $\boldsymbol{A}$ and $\boldsymbol{B}$ are the feature matrices of the negative teacher and the student, respectively, and $n$ is the number of elements of each matrix.

Due to the responses generating in the decoding phrase, we only conduct negative distillation on the intermediate layers of the decoder. For each decoder layer, the negative distillation objective of hidden states is defined as: 
\begin{equation}
    \label{eq:hid}
    \mathcal{L}_{hid}^l = \mathcal{L}_{MRSE}(\boldsymbol{H}_N^l, \boldsymbol{H}_S^l),
\end{equation}
where $\boldsymbol{H}_N^l$ and $\boldsymbol{H}_S^l$ are the output hidden states of the $l^{th}$ decode layer of $N$ and $S$, respectively. 

As the attention weights can learn substantial linguistic knowledge \citep{clark-etal-2019-bert}, it is beneficial for the student to further conduct negative distillation on the attention matrices, which is computed as follows:
\begin{equation}
\boldsymbol{A}=\frac{\boldsymbol{Q K}^{T}}{\sqrt{d_{k}}},
\end{equation}
\begin{equation}
\text { Attention }(\boldsymbol{Q}, \boldsymbol{K}, \boldsymbol{V})=\operatorname{softmax}(\boldsymbol{A}) \boldsymbol{V},
\end{equation}
where $\boldsymbol{Q}$, $\boldsymbol{K}$, and $\boldsymbol{V}$ are the matrices of queries, keys, and values, respectively, and $d_{k}$ is a scaling factor. 
Following \citet{jiao-etal-2020-tinybert}, the attention matrix $\boldsymbol{A}$ is chosen to calculate the distance rather than its softmax version $\operatorname{softmax}(\boldsymbol{A})$. 
Similar to Equation~\ref{eq:hid}, the negative distillation objective of attention matrices is formulated as:
\begin{equation}
\label{eq:att}
    \mathcal{L}_{att}^l = \mathcal{L}_{MRSE}(\boldsymbol{A}_N^l,\boldsymbol{A}_S^l)),
\end{equation}
where $\boldsymbol{A}_N^l$ and $\boldsymbol{A}_S^l$ are the attention matrices of the $l^{th}$ decoder layer of $N$ and $S$, respectively.

\renewcommand{\algorithmicrequire}{\textbf{Input:}}
\renewcommand{\algorithmicensure}{\textbf{Output:}}
\begin{algorithm}[h]
    \caption{Negative Distillation}
    \begin{algorithmic}[1]
    \REQUIRE
    $\mathcal{D}$: The raw training set; $H_{src}$ : The \textit{Source Entropy} filtering method; $N$ and $S$: The negative teacher and the student.
    \STATE \% Collection of negative training set.
    \STATE [Data\_entropy] $\leftarrow$ Calculate\_data\_entropy($\mathcal{D}$, $H_{src}$) using Eq.\ref{eq:h}
    \STATE Index\_list $\leftarrow$ Sort([Data\_entropy])
    \STATE $\mathcal{D}_N$ $\leftarrow$ Extract\_top\_data($\mathcal{D}$, Index\_list, 50\%)
    \STATE \% Training of negative teacher.
    \REPEAT
        \STATE Optimize $N$ by minimizing $\mathcal{L}_{mle}(N)$ on $\mathcal{D}_N$ using Eq.~\ref{eq:mle}
    \UNTIL{Convergence}
    \STATE \% Negative distillation.
    \REPEAT
        \STATE Optimize $S$ by minimizing $\mathcal{L}(S)$ on $\mathcal{D}$ using Eq.~\ref{eq:loss}
    \UNTIL{Convergence}
    \ENSURE
    $S$ : The trained student.

    \end{algorithmic}
    \label{algorithm}

\end{algorithm}

\subsection{Progressive Optimization}

The overall loss, combining the above negative distillation objectives and the MLE objective, is denoted as:

\begin{align}
\nonumber   \mathcal{L}= (1-\alpha)\mathcal{L}_{mle} + \\
\alpha(\mathcal{L}_{pred} +    \sum^l \mathcal{L}^l_{hid} + & \sum^l  \mathcal{L}^l_{att}),
\label{eq:loss}
\end{align}
where $\alpha$ is a hyper-parameter that balances the importance of supervised learning and negative distillation. 
For negative distillation, it would be better that the student has the ability to say something before it is reminded of what not to say. 
Thus, we perform a progressive distillation that first warms up the negative distillation ratio and then colds it down gradually. Inspired by the derivative of sigmoid function:
\begin{equation}
\sigma^{\prime}(z)=\sigma(z)(1-\sigma(z))=\frac{e^{-z}}{(e^{-z} + 1)^2},
\end{equation}
which shows a trend of gradual rise-fall, we define the balance coefficient $\alpha$ as:
\begin{equation}
\alpha = \lambda * \frac{e^{-z}}{(e^{-z} + 1)^2}, 
\label{eq:alpha}
\end{equation}
where $\lambda$ controls the peak value and $z$ is calculated by:
\begin{equation}
z(s) = \beta * (s - \gamma),
\end{equation}
where $s$ is the training step, and $\beta$ and $\gamma$ control the telescopic and translation transformation, respectively.

%% file: experiment.tex
\section{Experiments}

\subsection{Datasets}

In our experiments, two widely used dialogue datasets are employed to evaluate the proposed method: 
\textbf{DailyDialog}, which collects conversations that are similar to human daily communication \citep{li-etal-2017-dailydialog}, and \textbf{OpenSubtitles}, which consists of large-scale dialogues extracted from movie subtitles \citep{opensubtitles2009}.

In this work, we focus on the single-turn dialogue generation, thus we pre-process these two datasets into the (query, response) pairs. 
Table~\ref{tab:data_statistics} provides the statistics of both datasets.

\begin{table}[t]
\centering
\small

\begin{tabular}{lcccc}
\toprule
    Datasets        & Train & Valid & Test & Vocab\\ \midrule
    DailyDialog         & 68k & 6.8k & 6.8k  & 17,930\\
    OpenSubtitles        & 200k & 20k & 10k & 21,177  \\
\bottomrule
\end{tabular}
\caption{Statistics of two dialogue datasets in the experiments.}
\label{tab:data_statistics}
\end{table}

\begin{table*}[ht]
    \centering
    \small
    \begin{tabular}{l c c c c c c c c}
    \toprule
        Models & Dist-1 $\uparrow$ & Dist-2 $\uparrow$ & Dist-3 $\uparrow$ & LF $\uparrow$ & KL-1 $\downarrow$ & KL-2 $\downarrow$ & BLEU-3 $\uparrow$& BLEU-4 $\uparrow$\\  \midrule
        Standard & 0.0089 & 0.0313 & 0.0576 & 0.102 & 0.96 & \underline{0.53} & \textbf{0.384} & \textbf{0.395} \\
        NT & 0.0059 & 0.0293 & 0.0760 & 0.070 & 1.05 & 1.84 & 0.226 & 0.183 \\
        UL & 0.0062 & 0.0319 & \underline{0.0882} & 0.075 & 1.07 & 1.88 & 0.228 & 0.187 \\
        CVAE & 0.0024 & 0.0201 & 0.0821 & 0.029 & 1.54 & 2.02 & 0.144 & 0.087  \\
        FACE & \underline{0.0113} & \underline{0.0412} & 0.0763 & \underline{0.127} & \underline{0.83} & 0.57 & 0.099  & 0.063   \\

        ND & \textbf{0.0145} & \textbf{0.0678} & \textbf{0.1447} & \textbf{0.158} & \textbf{0.65} & \textbf{0.26} & \underline{0.381} & \underline{0.388}  \\

        \bottomrule
        \toprule
        Standard & \underline{0.0020} & 0.0071 & 0.0147 & 0.022   & 2.19 & \underline{1.40} & \textbf{0.355} & \underline{0.353}  \\
        NT & 0.0011 & 0.0045 & 0.0108 & 0.014  & 2.26 & 2.39 & 0.255 & 0.216 \\
        UL & 0.0015 & 0.0060 & 0.0151 & 0.018  & \textbf{1.85} & 1.97 & 0.303 & 0.269 \\
        CVAE & 0.0009 & 0.0055 & \underline{0.0182} & 0.013 & 2.76 & 2.70 & 0.134 & 0.084  \\
        FACE & \underline{0.0020} & \underline{0.0079} & 0.0166 & \underline{0.023} & \underline{2.03} & 1.56 & 0.353  & 0.339   \\

        ND & \textbf{0.0027} & \textbf{0.0102} & \textbf{0.0218} & \textbf{0.029}  & 2.10 & \textbf{1.23} & \textbf{0.355} & \textbf{0.355} \\

        \bottomrule
    \end{tabular}
    \caption{Automatic evaluation results using greedy search on DailyDialog (Up) and OpenSubtitles (Down). The best/second-best results are \textbf{bold}/\underline{underlined}. "$\uparrow$" means higher is better. "$\downarrow$" means lower is better.} 
    \label{tb:main}s
\end{table*}

\begin{table*}[ht]
    \centering
    \small
    \renewcommand\tabcolsep{2.0pt}
    \begin{tabular}{l|ccc|c|ccc|c|ccc|c}
     \toprule
      \multirow{2}{*}{vs. Models} & \multicolumn{3}{c|}{Informativeness} & \multirow{2}{*}{Kappa} & \multicolumn{3}{c|}{Relevance } & \multirow{2}{*}{Kappa} & \multicolumn{3}{c|}{Fluency} & \multirow{2}{*}{Kappa}  \\ 

      &  \makecell[c]{Win(\%)}  & \makecell[c]{Tie(\%)}  & \makecell[c]{Lose(\%)} & & \makecell[c]{Win(\%)}  & \makecell[c]{Tie(\%)}  & \makecell[c]{Lose(\%)} &  & \makecell[c]{Win(\%)}  & \makecell[c]{Tie(\%)}  & \makecell[c]{Lose(\%)}  & \\ \midrule
    Standard  & 77.3  & 18.7 & 4.0 & 0.456 & 48.0  & 34.7 & 17.3 & 0.453 & 14.7  & 76.0 & 9.3 & 0.491 \\ 
    NT & 31.3  & 38.7 & 30.0 & 0.669  & 54.7  & 32.0 & 13.3 & 0.421  & 91.3  & 8.0 & 0.7 & 0.497  \\
    UL & 44.7  & 28.7 & 26.7 & 0.411   & 66.0  & 23.3 & 10.7 & 0.425   & 92.7  & 7.3 & 0.0 & 0.614   \\
    \bottomrule
    \end{tabular}
    \caption{Results of human evaluations on DailyDialog. Our framework has a higher win rate than baselines.}
    \label{tb:human}
\end{table*}

\subsection{Experimental Settings}
We take the Transformer-based sequence-to-sequence model \citep{Transformer-Vaswani-2017} as the underlying model for all approaches. 

Following the settings of Transformer in \citet{csaky-etal-2019-improving}, both encoder and decoder contain 6 layers, in which the self-attention module has 8 attention heads and the number of feed-forward units is 2048. The size of hidden states is set to 512 and the dimension is 64 for query, key, and value. Please refer to Appendix~\ref{ap:de} for more details.

For the proposed approach, both the negative teacher network and the student network have the same settings in terms of the network architecture and hyper-parameters.
$\lambda$ in Equation~\ref{eq:alpha} is set to 4, making the peak value equal to 1. $\gamma$ is 25600 and $\beta$ is $6/\gamma$. For the temperature coefficient $t$, we simply set it to 1.

\subsection{Baselines}
We compare the proposed negative distillation (\textbf{ND}) approach with the standard Transformer, two existing negative training approaches and two extra diversity improving approaches:
\begin{itemize}
    \item \textbf{Standard} The vanilla Transformer-based sequence-to-sequence model with the MLE-based training (i.e., the cross-entropy based loss).
    \item \textbf{NT} (Negative Training) \citep{he-glass-2020-negative} During training, it first counts the frequency of all generated utterances and then conducts the negative update based on the high-frequency utterances.

    \item \textbf{UL} (Unlikelihood Training) \citep{li-etal-2020-dont} Different from \textbf{NT}, it calculates the frequency of all generated words instead of utterances and penalizes the high-frequency words by introducing an unlikelihood loss term.
    \item \textbf{CVAE} \citep{zhao-etal-2017-learning} A dialogue response generation model using conditional VAE to improve the diversity of generated responses.
    \item \textbf{FACE} \citep{DBLP:conf/www/JiangRMR19} It uses the frequency-aware cross-entropy loss to  tackle the low-diversity problem.

\end{itemize}

All the baselines are performed with the same architecture and hyper-parameters as ours. Following \citet{he-glass-2020-negative, li-etal-2020-dont}, we use greedy search as the decoding strategy for all baselines and our method. We also evaluate the performance with beam search (size 5) and obtain similar results (see \ref{exp:beam} for details). Details for baselines is describes in Appendix~\ref{ap:b}.

\subsection{Automatic Evaluation}
\paragraph{Metrics}
To evaluate whether negative distillation can effectively reduce the generic responses, we adopt \textbf{Dist-\{1,2,3\}} (distinct) \citep{li-etal-2016-diversity} to reflect the lexical diversity of the generated responses. It is a widely used metric that counts the proportion of unique unigrams/bigrams/trigrams. 
\textbf{LF} (low-frequency token ratio) \citep{DBLP:conf/iclr/LiWCUS0Z20} further measures the diversity of responses by calculating the ratio of low-frequency words in the generated responses. The threshold of low frequency is set to 100.

Besides, it is necessary to verify whether the models can ensure consistency while improving diversity.
So we use \textbf{KL-\{1,2\}} (KL divergence) \citep{csaky-etal-2019-improving}, which measures the distribution distance between the generated and the ground-truth responses, to reflect how well a model can approximate the ground-truth unigrams/bigrams distributions. \textbf{BLEU}  \citep{chen-cherry-2014-systematic} is also reported and it measures n-gram overlap between the generated and the ground-truth references.

\paragraph{Results}
Table~\ref{tb:main} shows the results obtained at the lowest point of the validation loss. 
We can see that our approach outperforms all baselines in diversity (\textbf{Dist} and \textbf{LF}) by a significant margin on both datasets, demonstrating that \textbf{ND} can effectively alleviate the generic response problem by using multi-level negative information. The \textbf{KL} and \textbf{BLEU} scores of \textbf{ND} are close to or better than \textbf{Standard}, which verifies that our method can maintain the consistency of responses while improving its diversity. 
To some extent, both \textbf{NT} and \textbf{UL} improve the diversity of words, especially for trigrams, but the low \textbf{LF} scores indicate that they reduce the high-frequency words but fail to increase the number of low-frequency's. What's worse, \textbf{BLEU} and \textbf{KL-2} scores of above two and \textbf{CVAE} sharply decline. It suggests that previous negative training approaches and other methods for diversity enhancement may harm the consistency and fluency of responses dramatically, which is not in line with the goals of the dialogue system. 
Our method obtains similar results with beam search. 
Please refer to \ref{exp:beam} for details.

\subsection{Human Evaluation}

Apart from automatic evaluations, we conduct human evaluations to further verify the effectiveness of our method than previous negative training methods. We randomly select 50 samples from the test set of DailyDialog, and three well-educated annotators are invited to judge which of the responses generated by \textbf{ND} and baselines is better (i.e., win, tie or loss) in terms of informativeness, relevance, and fluency. Informativeness reflects how much the information related to the query is contained in the generated response. Relevance reflects how likely the generated response is coherent to its query. Fluency reflects how likely the generated response is produced by human.

Table~\ref{tb:human} summarizes the human evaluation results. We can see that the proposed approach is overall better than all baselines. 
Specifically, \textbf{ND} achieves better performance than \textbf{Standard} in terms of informativeness and relevance, and remains competitive in fluency. 
Compared with both \textbf{NT} and \textbf{UL}, our approach shows significant advantages, especially in fluency. It indicates that their punishment for high-frequency tokens or utterances will lead to a serious non-fluency and inconsistency problem. 
We use Fleiss's kappa \citep{fleisskappa/measuring} to measure the inter-annotator agreement.

\subsection{Experimental Analysis}
We conduct extensive analysis on DailyDialog to investigate the effectiveness of the negative distillation in more details.

\paragraph{Ablation study} 
\begin{table}[th]
    \centering
    \small
    \begin{tabular}{l c c c c c c c c}
    \toprule
        Models & Dist-2 & Dist-3 & LF  & KL-2  & BLEU-4 \\  \midrule
        ND  & .0678 & .1447 & .158  & .26  & .388 \\
        w/o $\mathcal{L}_{pred}$  & .0529 & .1084 & .145  & .39  & .397 \\
        w/o $\mathcal{L}_{att}$  & .0517 & .1032 & .138  & .26 & .392 \\
        w/o $\mathcal{L}_{hid}$  & .0365 & .0677 & .109  & .62 & .380 \\ 
        w/o $\mathcal{L}_{neg}$  & .0313 & .0576 & .102  & .53 & .395 \\
        \bottomrule
    \end{tabular}
    \caption{Ablation studies of different negative distillation objectives in \textbf{ND}.}
    \label{tb:ablation}
\end{table}

We study the effects of different negative distillation objectives by ablating the prediction layer distillation (w/o $\mathcal{L}_{pred}$), the attention distillation (w/o $\mathcal{L}_{att}$), the hidden state distillation (w/o $\mathcal{L}_{hid}$), and the whole negative distillation (w/o $\mathcal{L}_{neg}$, i.e. Standard). 
The results in Table~\ref{tb:ablation} show that all three proposed negative distillation objectives are useful for improving the diversity. 
The significant decline in w/o $\mathcal{L}_{hid}$ indicates that the negative information in intermediate layers is very important for \textbf{ND}. 
w/o $\mathcal{L}_{att}$ is better than w/o $\mathcal{L}_{hid}$, attributing to the more abundant information in hidden states.

\paragraph{Does \textit{source entropy} work?}
To verify whether the \textit{source entropy} filtering method can collect the generic responses, we select the top 50\% and the bottom 50\% of the sorted training set as $\mathcal{D}_t$ and $\mathcal{D}_b$, respectively. Then we train $N_t$ and $N_b$ on the corresponding sub-sets. 
From Table~\ref{exp:ent}, we can see that $N_b$ outperforms $N_t$ in all the diversity-related metrics, indicating the effectiveness of \textit{source entropy}.

\begin{table}[th]
    \centering
    \small
    \begin{tabular}{l c c c c}
    \toprule
        Models & Dist-1 & Dist-2 & Dist-3 & LF \\  \midrule
        $N_t$  & 0.0024  & 0.0078  & 0.0134  & 0.0331  \\
        $N_b$  & \textbf{0.0040}  & \textbf{0.0121}  & \textbf{0.0215}  & \textbf{0.0444}   \\
        \bottomrule
    \end{tabular}
    \caption{Effect of the \textit{source entropy} filtering method.}
    \label{exp:ent}
\end{table}

\paragraph{Can the negative knowledge be transferred?}
We take $N_t$ and $N_b$ as the negative teachers for the students $S_t$ and $S_b$, respectively. Then we conduct negative distillation on both $S_t$ and $S_b$. 
The results in Table~\ref{tb:nt} demonstrate that $S_t$ obtains more gains in diversity than $S_b$, indicating $S_t$ gets rid of more negative knowledge. It can be further verified by the results of 

\begin{table}[th]
    \centering
    \small
    \begin{tabular}{l c c c c c c c c }
    \toprule
        Models & Dist-2 & Dist-3 & LF & KL-2 & BLEU-4\\  \midrule
      
        ${S}_t$  & \textbf{0.0678} & \textbf{0.1447} & \textbf{0.158} & \textbf{0.26} & \textbf{0.388} \\
        ${S}_b$  & 0.0409  & 0.0844  & 0.097 & 0.40 & 0.386  \\     
        \bottomrule
    \end{tabular}
    \caption{Effect of negative knowledge.}
    \label{tb:nt}
\end{table}

\begin{table}[th]
    \centering
    \small
    \begin{tabular}{l c c c c c  }
    \toprule
        Models & Dist-2 & Dist-3 & LF & KL-2 & BLEU-4\\  \midrule
        Standard & .0313 & .0576 & .102  & .53  & .395 \\    
        ND (fixed $\alpha$)  & .0392 & .0793  & .123 & .42 & .386  \\
        ND  & .0678 & .1447 & .158  & .26& .388  \\
    \bottomrule
    \end{tabular}
    \caption{Effect of progressive distillation.}
    \label{tb:alpha}
\end{table}

\begin{table}[th]
    \centering
    \small
    \begin{tabular}{l c c c c c c c c }
    \toprule
        Models & Dist-1 & Dist-2 & Dist-3 & LF \\  \midrule
        ND (random target) & 0.0040 & 0.0109 & 0.0170 & 0.053 \\    
        ND (hard target)  & 0.0136  & 0.0620 & 0.1344  & 0.139 \\
        ND (soft target) & 0.0145 & 0.0678 & 0.1447 & 0.158  \\
        \bottomrule
    \end{tabular}

    \caption{Comparison of soft targets, hard targets, and random targets for negative distillation.}
    \label{tb:soft}
\end{table}

\paragraph{Study of soft target}
To evaluate the superiority of soft targets for negative distillation, we sample responses (i.e., hard target) by greedy search on the predictions of negative teachers for comparison. The results in Table~\ref{tb:soft} show that ND with soft targets can diversify the responses more effectively, demonstrating the advantages of richer negative information (e.g., the similarity between labels) in soft targets. 
What's more, we randomly select responses from the negative training set $\mathcal{D}_N$ as negative targets. The sharp decline in performance proves that the negative teacher can produce targeted generic responses.

\begin{table*}[ht]
    \centering
    \small
    \begin{tabular}{l c c c c c c c c}
    \toprule
        Models & Dist-1 $\uparrow$ & Dist-2 $\uparrow$ & Dist-3 $\uparrow$ & LF $\uparrow$ & KL-1 $\downarrow$ & KL-2 $\downarrow$ & BLEU-3 $\uparrow$& BLEU-4 $\uparrow$\\  \midrule
        Standard & 0.0060 & 0.0238 & 0.0455 & 0.068 & 0.92 & 0.62 & 0.375 & 0.372 \\
        NT & 0.0077 & 0.0326 & 0.0640 & 0.083 & 0.76 & 0.70 & 0.349 & 0.334 \\
        UL & 0.0059 & 0.0270 & 0.0570  & 0.075 & 0.82 & 0.95 & 0.321 & 0.294 \\
        CVAE & 0.0022 & 0.0152 & 0.0579  & 0.026 & 1.60 & 1.94 & 0.144 & 0.087  \\
        FACE & 0.0074 & 0.0305 & 0.0591 & 0.086 & 0.73 & 0.64 & 0.123   & 0.077\\
        ND & \textbf{0.0108} & \textbf{0.0427} & \textbf{0.0799} & \textbf{0.120} & \textbf{0.72} & \textbf{0.05} & \textbf{0.386} & \textbf{0.404} \\

        \bottomrule

    \end{tabular}
    \caption{Automatic evaluation results using beam search (size 5) on DailyDialog. Our approach (\textbf{ND}) outperform baselines on all the metrics.}
    \label{tb:beam}
\end{table*}

\paragraph{Effect of progressive distillation}
In order to verify the effectiveness of progressive negative distillation, we conduct negative distillation with fixed $\alpha$. 
The value is obtained by calculating the average of $\alpha$ in Equation~\ref{eq:alpha} across the convergence steps. 

The results in Table~\ref{tb:alpha} demonstrate that the progressive distillation policy can help the student exploit negative knowledge more effectively. Besides, note that ND with fixed $\alpha$ also outperforms the Standard model.

\paragraph{Evaluation results with beam search}
\label{exp:beam}

\citet{he-glass-2020-negative} and \citet{li-etal-2020-dont} choose greedy decoding due to its simplicity and higher diversity than beam decoding. However, we find that both \textbf{NT} and \textbf{UL} tend to generate long but non-fluent and incoherent responses. So we conduct beam search with adding the length penalty. Table~\ref{tb:beam} summarizes the results and it shows that both two baselines get better \textbf{KL} and \textbf{BLEU} scores than using greedy search due to shorter responses. \textbf{ND} outperform baselines on all the metrics, confirming the effectiveness of our method.

\paragraph{Case study}

\begin{table}[th]
\centering

\small
\begin{tabular}{@{}p{\linewidth}@{}}
\toprule
\textbf{Input:} The phone turns off by itself.\\
\midrule
\textbf{Standard:} What's the problem? \\
\textbf{NT:} I see. Is a part of the most important things. And I'll be happy to keep you there. Is the most important thing for you... \\
\textbf{UL:} I see. Is not allowed to be a good thing. Birthday. Birthday. - hour are very important...... \\

\textbf{ND:} I am calling to tell you that he is in the right desk.  \\
\midrule
\midrule
\textbf{Input:} Well, I bought this skirt here yesterday afternoon. But I got the wrong size. It's too small for my daughter. I wonder if I could refund it.\\
\midrule
\textbf{Standard:} That's fine. \\
\textbf{NT:} That's fine.  Your mind and address number. number. And date are available. And so on. Is very simple...... \\
\textbf{UL:} I'm sorry, sir. But you've got to work overtime before you leave the contract. Service is very important. Service. Service usually be late.\\
\textbf{ND:} I think you have to pay the money. \\
\bottomrule
\end{tabular}
\caption{Examples of generated responses.}

\label{tb:case study}

\end{table}

Table~\ref{tb:case study} shows some cases generated by the proposed method and baselines. \textbf{Standard} prefers generic and meaningless responses. Both \textbf{NT} and \textbf{UL} tend to generate a short generic sentence followed by a incoherent and non-fluent subsequence. In contrast, \textbf{ND} can produce diverse and coherent responses.

%% file: related_work.tex
\section{Related work}
\paragraph{Diversity Dialogue Learning}
There are two lines of work for solving the generic response problem:
One line promotes the diversity from positive view, which is outside of our work. Specially, previous work includes MMI \citep{li-etal-2016-diversity}, GAN \citep{li-etal-2017-adversarial,DBLP:conf/nips/ZhangGGGLBD18}, CVAE \citep{zhao-etal-2017-learning}, BT \citep{su-etal-2020-diversifying}, FACE \citep{DBLP:conf/www/JiangRMR19}, AdaLabel \citep{wang-etal-2021-diversifying}, IAT \citep{zhou-etal-2021-learning}, and Nucleus Sampling \citep{DBLP:conf/iclr/HoltzmanBDFC20}.  
The other line alleviates the generic response problem using negative training. \citet{he-glass-2020-negative} regards frequent response problem as a sub-problem of the generic response problem and conduct negative update for the high-frequency responses during training. \citet{li-etal-2020-dont} focuses on high-frequency tokens rather than tokens and punishes them by using the unlikelihood objective \citep{DBLP:conf/iclr/WelleckKRDCW20}. Both of them handle the generic response problem only from the angle of reducing frequency, thus can not capture all the features of generic replies.

\paragraph{Negative Training for Dialogue Learning}
Negative training for retrieval-based dialogue learning has been previously extensively studied \citep{DBLP:conf/iclr/HumeauSLW20, DBLP:conf/percom/NugmanovaSLC19}, while we focus on the dialogue generation in this work. \citet{he-glass-2020-negative} uses negative training to prevent generic and malicious responses in dialogue models. \citet{li-etal-2020-dont} generalizes unlikelihood to dialogue generation for improving repetition, specificity and coherence. \citet{lagutin-etal-2021-implicit} proposes implicit unlikelihood training to minimize repetition.
Our work proposes a new negative training paradigm aimed at improving the diversity of dialogue responses while avoiding the problem of poor consistency and fluency of previous work.

%% file: conclusion.tex
\section{Conclusion}
We present a novel negative training paradigm to improve the diversity of dialogue responses. 
It formulates the conventional negative training as a knowledge distillation process, which is rarely explored before. 
The negative teacher can produce the corresponding generic and dull responses given any query, which naturally avoids problems that hinder previous negative training methods. 
Besides, we further boost the performance of negative distillation by exploiting richer information, i.e., multi-level features.
Extensive experiments validate the superiority of our proposed method compared with prior negative training work.

A limitation of our work is that we only focus on the generic response problem. For future work, we will extend the proposed negative distillation to handle other generation problems, such as inconsistency and lacking personas or emotions. 

%% file: appendix.tex
\appendix 
\section{Details for Implementations}
\label{ap:de}
Here are some implementation details of our experiments. Dropout \citep{DBLP:journals/jmlr/SrivastavaHKSS14} is used for the self-attention module, the feed-forward layer, and the activation layer, and the rate of all three is set to 0.1. We also use label smoothing \citep{DBLP:conf/cvpr/SzegedyVISW16} and the smoothing value is 0.1. 
The batch size is set to 256. We use the Adam optimizer \citep{DBLP:journals/corr/KingmaB14} and employ the \textit{warm-up} \citep{DBLP:conf/cvpr/HeZRS16} trick to adjust the learning rate during training. The warm-up steps $s_\text{wp}$ are 128k and 256k for DailyDialog and OpenSubtitles, respectively. The learning rate is computed as follows:
\begin{equation}
lr = \frac{2 \cdot \min(\frac{1}{\sqrt{s}}, \frac{s}{\sqrt{s_\text{wp}^{3}}})}{\sqrt{d_\text{model}}},
\end{equation}
where $lr$ is the learning rate at the $s^{th}$ step of training and $d_\text{model}$ is the size of hidden states. 
We implement all approaches with Pytorch 1.7, and conduct all experiments on RTX 3090.

\section{Baselines}
\label{ap:b}
For \textbf{NT}, the threshold $r_{\text{thres}}$ is set to 1\% and the weight coefficient $\lambda_{\text{POS}}$ is set to 1 as the authors' suggestion. 
For \textbf{UL}, we search the mixing hyper-parameter $\alpha$ in $[1, 10, 100, 1000]$ and 1000 is selected for its best performance. Both \textbf{NT} and \textbf{UL} are refined on the well-trained \textbf{Standard} model. For \textbf{CAVE}, we set the latent size with patience to 256 and 64 for DailyDialog and OpenSubtitles, respectively. And for \textbf{FACE}, we use the "output frequency" and "pre-weight" version as the author suggested.

We also compare the proposed method (ND) with AdaLabel \citep{wang-etal-2021-diversifying}, although AdaLabel alleviates the generic response problem from the perspective of target regularization rather than negative training. The results in Table~\ref{tb:adalabel} confirms the superior performance of our method for improving the diversity of generated responses. In addition, the negative distillation method can be readily extended to other generation problem, while AdaLabel mainly focuses on diversity.

\begin{table*}[ht]
    \centering
    \small
    \begin{tabular}{l c c c c c c c c}
    \toprule
        Models & Dist-1 $\uparrow$ & Dist-2 $\uparrow$ & Dist-3 $\uparrow$ & LF $\uparrow$ & KL-1 $\downarrow$ & KL-2 $\downarrow$ & BLEU-3 $\uparrow$& BLEU-4 $\uparrow$\\  \midrule
        AdaLabel & 0.0100 & 0.0397 & 0.0757 & 0.105 & 0.89 & 0.59 & 0.097 & 0.061  \\
        ND & 0.0145 & 0.0678 & 0.1447 & 0.158 & 0.65 & 0.26 & 0.381 & 0.388  \\
        \bottomrule
        \toprule
        AdaLabel & 0.0065 & 0.0259 & 0.0476 & 0.064 & 1.11 & 0.83 & 0.109 & 0.069  \\
        ND & 0.0108 & 0.0427 & 0.0799 & 0.120 & 0.72 & 0.05 & 0.386 & 0.404 \\
        \bottomrule
    \end{tabular}
    \caption{Comparing with AdaLabel by greedy search(Up) and beam search(Down) on DailyDialog. "$\uparrow$" means higher is better. "$\downarrow$" means lower is better.} 
    \label{tb:adalabel}
\end{table*}